\documentclass[twocolumn]{article}
\usepackage[utf8]{inputenc}

\usepackage[sorting=none,url=false]{biblatex}
\usepackage{amsmath}

\usepackage{xcolor}

\usepackage{graphicx}

\usepackage{blindtext}

\usepackage{amssymb}

\usepackage{algorithm}
\usepackage{algorithmicx}
\usepackage{algpseudocode}

\makeatletter
\renewcommand*{\ALG@name}{Procedure}
\makeatother

\begin{document}

\onecolumn

\begin{center}
{\Large\bf Adaptive Modeling Powers Fast Multi-parameter Fitting of CARS Spectra}\\
\vspace{.25in}
Gregory J. Hunt$^1$, Cody R. Ground$^2$, Andrew D. Cutler$^3$\\\vspace{.25in}

\end{center}

$^1$ Department of Mathematics, William \& Mary, PO Box 8763 Williamsburg, VA 23187, United States of America\\

$^2$ Hypersonic Airbreathing Propulsion Branch, NASA Langley Research Center, 12 Langley Blvd. Hampton, VA 23681, United States of America\\

$^3$ The George Washington University, Mechanical and Aerospace Engineering, 800 22nd St NW, Washington, DC 20052, USA.\\

\begin{center}
\today
\vspace{.5in}
\end{center}

\begin{abstract}
    Coherent anti-Stokes Raman Spectroscopy (CARS) is a laser-based measurement technique widely applied across many science and engineering disciplines to perform non-intrusive gas diagnostics. CARS is often used to study combustion, where the measured spectra can be used to simultaneously recover multiple flow parameters from the reacting gas such as temperature and relative species mole fractions. This is typically done by using numerical optimization to find the flow parameters for which a theoretical model of the CARS spectra best matches the actual measurements. The most commonly used theoretical model is the CARSFT spectrum calculator. Unfortunately, this CARSFT spectrum generator is computationally expensive and using it to recover multiple flow parameters can be prohibitively time-consuming, especially when experiments have hundreds or thousands of measurements distributed over time or space. To overcome these issues, several methods have been developed to approximate CARSFT using a library of pre-computed theoretical spectra. In this work we present a new approach that leverages ideas from the machine learning literature to build an adaptively smoothed kernel-based approximator. In application on a simulated dual-pump CARS experiment probing a $H_2/$air flame, we show that the approach can use a small number library spectra to quickly and accurately recover temperature and four gas species' mole fractions. The method's flexibility allows fine-tuned navigation of the trade-off between speed and accuracy, and makes the approach suitable for a wide range of problems and flow regimes.  
\end{abstract}

\twocolumn

\section{Introduction}\label{sec:intro}

Coherent anti-Stokes Raman Spectroscopy (CARS) is a spectroscopic technique with application across a wide range of scientific fields including physics, chemistry, biology, engineering, and atmospheric science \parencite{Tolles:77,Danehy2003,greif2021, Evans16807,evans2008}. In gas diagnostics and combustion, CARS is commonly used to estimate flow parameters like temperature and relative species mole fractions \parencite{Danehy2003,greif2021}. This is possible as the shape of the CARS spectra depends on these underlying flow parameters. 

\begin{figure}[!t]
    \centering
    \includegraphics[width=.45\textwidth]{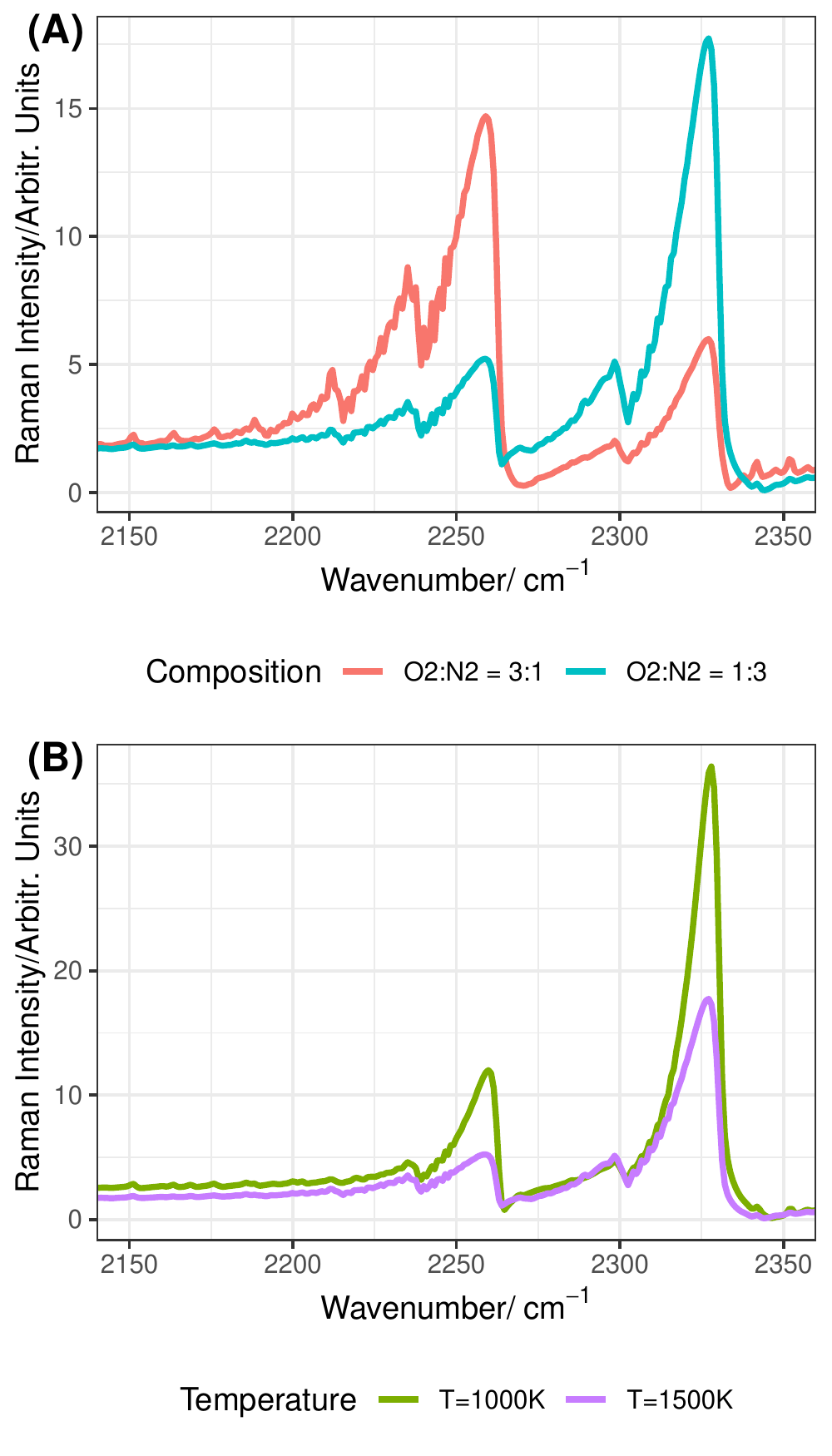}
    \caption{(A) Example dual-pump CARS spectra at $T=1500K$. The blue curve is for a gas composed of $O_2$ and $N_2$ in a 3:1 ratio. Red curve is for a situation where $T=1500K$ and a 1:3 ratio of $O_2$ to $N_2$. (B) Similar example of dual-pump CARS spectra holding the $O_2:N_2$ ratio fixed at 1:3 but varying the temperature between $1000K$ and $1500K$. The instrument parameters match those of Ref. \cite{obyrne2007}}
    \label{fig:ex_spectra}
\end{figure}

Figure~\ref{fig:ex_spectra} displays spectra for several different combinations of flow parameters. In subplot (A) we show two example spectra for differing relative mole fractions of $N_2$ and $O_2$ and in subplot~(B) we show two spectra with the same species mixtures but different gas temperatures. The dependence of the spectra on these flow parameters like temperature and relative mole fraction means one can estimate flow parameters by finding the values for which a theoretical model of the spectrum best matches the actual measurements. Typically this is done by minimizing the squared difference between theory and measurement using numerical optimization. 

The most popular theoretical model used to generate CARS spectra is the CARSFT spectrum calculator developed by Sandia National Laboratory \parencite{carsft}. While accurate, generating spectra using the CARSFT spectrum calculator is computationally expensive. This can make parameter estimation slow as optimization routines may need to calculate the theoretical spectra hundreds or thousands of times to fit a single spectrum. This problem is often further compounded when estimating flow properties of multiple spectra---such as in experiments where spectra are measured at high sampling rates and/or at multiple locations in space \parencite{VanderSteen1997, obyrne2007, Dedic17}. In either one of these instances, one may have hundreds or thousands of measured spectra to fit. 

The slow nature of fitting CARS spectra has been widely recognized in the literature and consequently several approaches have been employed to tackle this problem. The most common way to speed up fitting CARS spectra is to replace the slow theoretical model with a faster approximation \parencite{cutler2011}. This approach is simliar to surrogate modeling in other areas of engineering \parencite{QUEIPO20051}. For CARS, this surrogate model is typically constructed using a library of theoretical spectra generated at known flow parameters. The goal is to use this library to approximate the relationship between flow parameters and their corresponding spectra. 
The most common approach is to fit multivariate interpolating Lagrange polynomials \parencite{Danehy2003,obyrne2007,cutler2011,cutler2015}.

A significant issue with Lagrange interpolation is that the method can only interpolate over a regular grid of parameters. This has several knock-on effects. Firstly, it does not scale well when applied to complicated multi-parameter problems as the size of such a grid grows exponentially in the number of parameters. This is particularly problematic in multispecies CARS techniques, such as dual-pump CARS, that target several resonant species where the spectra can be used to simultaneously recover the relative mole fractions of the multiple resonant species in addition to the gas temperature \parencite{HanCock97}. For example, if we construct a library using ten different values for each of five parameters this will require $10^5 = 100,000$ library spectra. Fine-tuned control over library size is also difficult as one cannot add or remove arbitrary spectra from the library since this may not form a regular grid. Furthermore, the physically meaningful space of flow parameters is not a grid. For example, the mole fractions of the resonant gas species live in a simplex since they must be positive and sum to one or less. More broadly, the physically realizable parameter space is further constrained by the conservation of mass, momentum, and energy, the laws of thermodynamics, and the boundary conditions for the problem. Thus, the physically meaningful parameter space is often irregularly shaped. One may partially side-step this problem by interpolating over a regular grid in a small rectangle encompassing the physically meaningful space. Even then many of the spectra in the library will likely correspond to flow conditions that are never actually encountered or, furthermore, to conditions that are completely nonphysical. For example, library spectra corresponding to the nonphysical situation where the sum of the mole fractions of the resonant and non-resonant species are greater than 1 could be required in such a regularly-gridded library \cite{cutler2011}. Often the minimally sized bounding rectangle is still significantly larger than the physically meaningful parameter space. For example, with $N$ parameters a simplex has a volume $N!$ times smaller than its bounding rectangle. This difference in size can become quite significant for high-dimensional problems. Consequently, even for the smallest possible bounding rectangle, the regularly-gridded parameter space required for Lagrange interpolation will wastefully increase computational complexity for little benefit since the interpolated fit is never actually evaluated at the unrealized and nonphysical points.

These problems with Lagrange interpolation often lead to oversized libraries that are both slow to construct and yield interpolators that are slow to evaluate. Consequently, a small number of alternative approaches have been developed to improve estimation speed. A popular approach, formulated by Cutler~et~al.,~facilitates fast recovery of multiple gas species mole fractions using Lagrange interpolation on a small ``sparse'' library generated by CARSFT that is augmented by newly calculated theoretical spectra generated during the solution process \parencite{cutler2011,cutler2015}. Alternatively, Greifenstein~et~al.~propose a method to enable fast parameter recovery by compressing the computational footprint of large libraries using properties of the experimental apparatus function \parencite{greif2021}. The method currently only supports single species nitrogen rovibrational CARS. 

In this paper we consider an alternative approach and investigate how CARS spectra might be quickly fit by applying a flexible kernel-based method. The approach we present is inspired by previous work investigating the approximation of Rayleigh-Brillouin scattering \parencite{Hunt2020}. However, more broadly, this kernel-based method is part of a class of approaches have proven successful in other approximation problems widely across scientific fields \parencite{QUEIPO20051,xiao,Zhang2012,Drucker1997,Hastie2009}. 

The remainder of the paper is structured as follows. In Section~\ref{sec:method} we briefly describe the class of methods we use. Subsequently, in Section~\ref{sec:eval} we investigate performance of such models in approximating CARS spectra. Here, we show that they are fast, accurate, and adaptable. We demonstrate this efficacy in the context of a five-parameter problem recovering temperature and mole fractions of four gas species in a theoretical dual-pump CARS experiment of $H_2/$air combustion. Finally, in Section~\ref{sec:conclusion} we conclude. 

\section{Flexible Spectral Modeling}\label{sec:method}

Our goal is to estimate flow parameters from measured spectra. Assume we wish to recover $P$ flow parameters generically denoted by the vector $x=(x_1,\ldots,x_P)$. In section~\ref{sec:eval} we will demonstrate the following method in a theoretical dual-pump CARS experiment with $P=5$ parameters: temperature and the mole fractions of $N_2$, $H_2$, $O_2$, and $H_2O$, respectively. However, this method is generally extensible to any CARS experiment, targeting either vibrational or rotational transitions, with any number of parameters $P$. This includes the case where $P=1$, as in a traditional vibrational $N_2$ CARS experiment, where the measured spectra are used solely to recover a gas temperature measurement. For a given set of parameters $x$ let $r(x) = (r_1(x),\ldots,r_M(x))$ denote the theoretical spectral intensity observed at $M$ fixed wavenumbers. In our application $r(x)$ will be generated by CARSFT. This $r(x)$ is precisely what is displayed in Figure~\ref{fig:ex_spectra}.

To make parameter estimation quick, we want to replace the slow theoretical model $r$ with a faster approximation denoted $\hat{r}$. Library-based approaches learn this approximation on the basis of a library of $N$ chosen flow parameters and their corresponding theoretical spectra generated by $r$. Let our library of $N$ spectra be denoted by the collection of vectors $r^{(1)},\ldots,r^{(N)}\in\mathbb{R}^M$ each corresponding to a library parameter $x^{(1)},\ldots,x^{(N)}\in\mathbb{R}^P$ so that $r^{(n)} = r\left(x^{(n)}\right)$. 


Given a new $x\in\mathbb{R}^P$ we want to approximate $r(x)=(r_1(x),\ldots,r_M(x))$ using this library. To do this we approximate our spectra as a linear combination of a basis of kernel functions \parencite{Cucker2001}. In particular, we approximate the $m^{th}$ component function $r_m$ as $\hat{r}_m$ so that 
\begin{equation}\label{eqn:ksum}
r_m(x) \approx \hat{r}_m(x) = \sum_{n=1}^N w_{mn}k_\gamma\left(x,x^{(n)}\right)
\end{equation}
where the $w_{mn}$ are real-valued weights we will learn and $k_\gamma$ is a kernel function that forms the basis of the approximation expansion. In our application we use a radially symmetric Gaussian kernel $k_\gamma$ so that $k_\gamma(x,y) = \exp\left(-\gamma||x-y||^2\right)$. We choose this class of Gaussian kernel functions because they are flexible and generally perform well, however there are other possible choices (c.f. \parencite{Hastie2009}). The parameter $\gamma$ in Equation~(\ref{eqn:ksum}) is a tuning parameter that controls the smoothness of the model and will be discussed in detail in Section~\ref{sec:xv}. Carefully tuning this parameter $\gamma$ will be crucial to a successful approximation. 

While Equation~(\ref{eqn:ksum}) displays the form for each component approximator $\hat{r}_m$, all of these $\hat{r}_m$ may be combined together into a succinct closed-form estimator for $r$. To do this, let $w_m = (w_{m1},\ldots,w_{mN})\in\mathbb{R}^N$ be our vector of weights for the $m^{th}$ component function and $k_\gamma(x) = (k_\gamma(x,x^{(1)}),\ldots,k_\gamma(x,x^{(N)})) \in\mathbb{R}^N$ be the vector of kernel functions. With this notation, we can write the approximator for $r_m$ as $\hat{r}_m(x) = w_m^Tk_\gamma(x)$. Furthermore, if $W$ is the $M \times N$ matrix whose $m^{th}$ row is $w_m$ so that
\[
W = \begin{bmatrix}
$---$& w_1& $---$ \\
$---$& w_2& $---$\\
& \vdots& \\
$---$& w_M& $---$
\end{bmatrix}
\]
 then we may write the entire expression for $\hat{r}$ as
\begin{equation}\label{eqn:approx}
    \hat{r}(x) = W k_\gamma(x).
\end{equation}

Concretely, for our Gaussian kernel this approximator takes on a exponential form
\begin{equation}\label{eqn:kmod}
\hat{r}(x)=W\exp\left(-\gamma||x-X||^2\right)
\end{equation}
where $X$ is our $P\times N$ matrix of library parameters, the notation ``$||x-X||^2$'' is short-hand for the $N\times 1$ vector whose elements are $||x-x^{(n)}||^2$ and the exponentiation ``$\exp$'' is applied element-wise.

The power of such kernel-based approximators in a multi-parameter setting stems from their ability to flexibly model functions of multiple variables. While previous interpolation approaches have modeled $r$ using multivariate polynomial interpolation on a grid, this kernel-based approach employs a grid-free approach to effectively find a model for $r$ by searching among all possible multivariate polynomials of any degree. This approach uses what is known as the ``kernel trick'' which allows the approximator to implicitly exploit the Taylor series expansion of the exponential function to find a fit among the collection of all polynomials without ever actually having to do computations in this intractable space \parencite{Hastie2009}. This yields a very flexible and powerful class of estimators with the simple exponential form seen in Equation~(\ref{eqn:kmod}).

To calculate this efficiently, one may expand out the exponentiated term $||x-X||^2$ as 
\begin{equation}\label{eqn:exp}
x^Tx - 2X^Tx + diag^{-1}\left(X^TX\right)
\end{equation}
where $diag^{-1}$ retrieves the diagonal elements of $X^TX$ and the addition/subtraction is element-wise. Since $W$ is determined in advance and $X$ (and $diag^{-1}\left(X^TX\right)$) are known, then the expensive operations of calculating  $\hat{r}$ are (1) calculating $X^Tx$, (2) exponentiating the term in Equation~(\ref{eqn:exp}), and (3) multiplying this by $W$. Consequently, predictions with this model can be efficiently implemented in code using just two matrix multiplications and exponentiation. These operations are very efficient in any modern scientific computing language. 

Ultimately, to calculate $\hat{r}$ we need to determine values for $W$ and $\gamma$ to make the approximation accurate. Section~\ref{sec:w} will discuss how we estimate $W$ while Section~\ref{sec:xv} will discuss $\gamma$. This latter parameter $\gamma$ will control the adaptability of this kernel-based approach by controlling the smoothness of the fit. This parameter does this by modulating the relative influence of the library parameters on $\hat{r}$. Large values of $\gamma$ means $\hat{r}(x)$ is strongly influenced by parameters in the library close to $x$ and consequently $\hat{r}$ will take on a complex non-linear fit. This can be good as it means the approximator can correctly model potentially complex $r$ but makes the method susceptible to over-fitting. Conversely, a small $\gamma$ produces a less complex smoother fit. Such a fit is less adaptable to highly non-linear shapes but also less likely to over-fit. This parameter $\gamma$ will be important for adapting this kernel-based approach to CARS. By carefully tuning $\gamma$ we will show that we can adaptively build a quick and accurate surrogate model.

Finally, an important feature of our proposed approach is that it does not require a library that forms a regular grid. Indeed, we can fit this approximator using libraries of arbitrary structure. This makes the approach highly adaptable. For example, one could build up the library one spectra at a time to tune the speed-accuracy trade-off in an adaptive manner. Furthermore, one need only include spectra in the library that are physically meaningful. For example, unlike Lagrange interpolated methods, we need only have library spectra whose mole fractions are relevant to the combustion problem of interest and physically meaningful. This helps minimize library size which reduces computational complexity and increases fitting speed. 


\subsection{Learning $W$}\label{sec:w}

There are many ways one might find good values for $W$ e.g. using a kernel regression or support vector regression approach \parencite{Hastie2009,Drucker1997}. Since the underlying $r$ does not to have a random error component we estimate $W$ by imposing an interpolation condition on $\hat{r}$ so that our approximation is exact for spectra in our library. That is $\hat{r}$ and $r$ match exactly for spectra in our library so that \[
r\left(x^{(n)}\right) = \hat{r}\left(x^{(n)}\right).
\] 
This interpolation condition yields a model that is closely related to radial basis function (RBF) interpolation \parencite{meshfree}. 

To use this condition to actually form an estimate for $W$ we let $R$ be the $M \times N$ matrix whose columns are the library spectra so that
\[
R = \begin{bmatrix}
\vert& & \vert\\
r\left(x^{(1)}\right)&  \cdots&  r\left(x^{(N)}\right)\\
\vert& & \vert\\
\end{bmatrix}
\] 
and let $K$ be the matrix whose columns are the $k_\gamma\left(x^{(n)}\right)$ so that 
\[
K = \begin{bmatrix}
\vert& & \vert\\
k_\gamma\left(x^{(1)}\right)&  \cdots&  k_\gamma\left(x^{(N)}\right)\\
\vert& & \vert\\
\end{bmatrix}.
\]
Then, given these matrices, our interpolation condition is
\[
R = WK
\]
and so we estimate W as 
\[
W = RK^{-1}. 
\]
Note that we are always guaranteed that $K$ is invertible so long as the library parameters $x^{(n)}$ are all unique \parencite{mairhuber}.

\subsection{Adaptively Smoothing with $\gamma$}\label{sec:xv}

\begin{figure*}[!t]
    \centering
    \includegraphics[width=\textwidth]{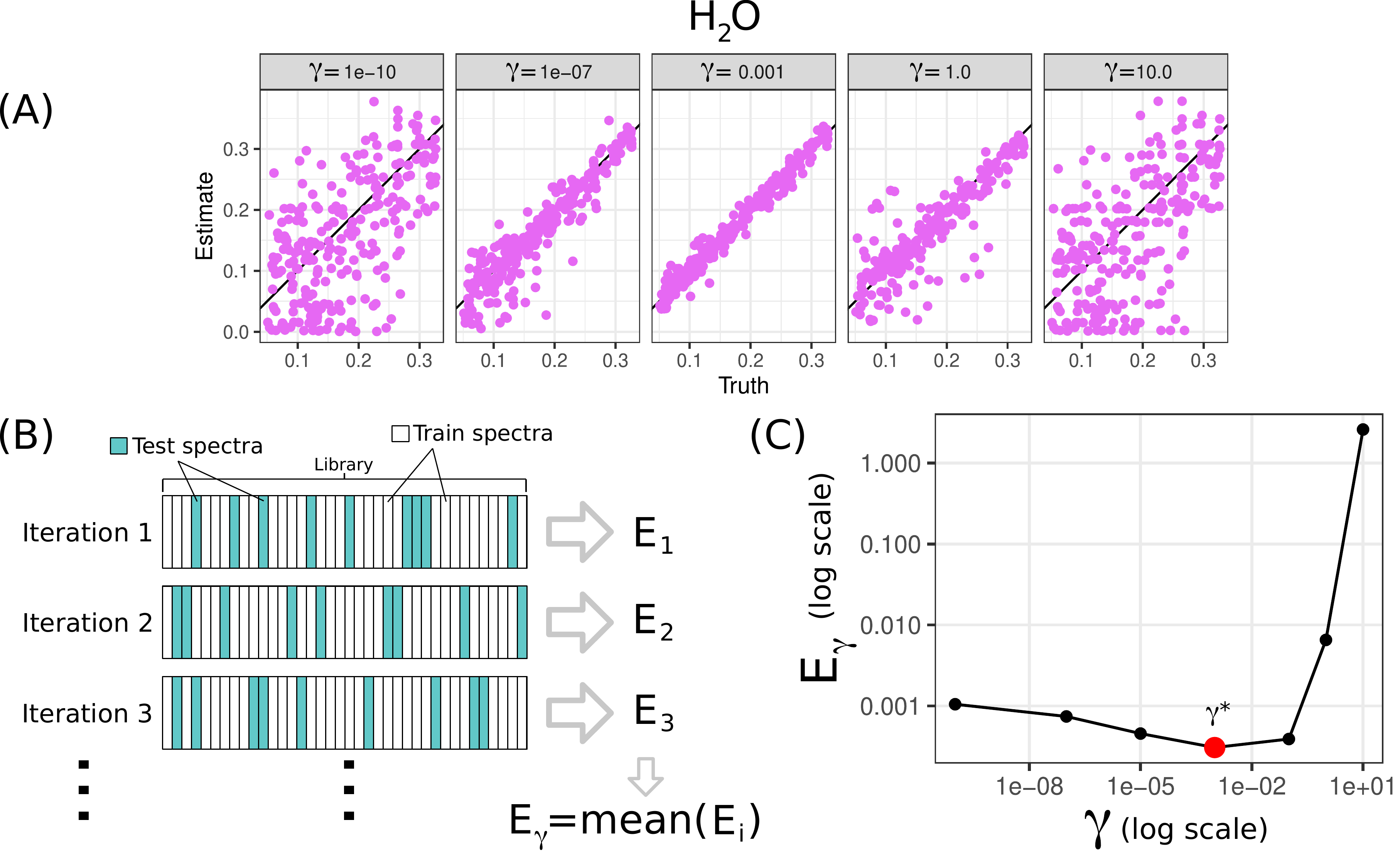}
    \caption{(A) Scatter plots of truth v. estimates for recovering $H_2O$ mole fraction from the validation spectra using $N=200$ library spectra and adding noise to the validation spectra with a signal-to-noise ratio of $50$. This scatter plot is replicated for several values of $\gamma$. From this we can see that accuracy depends heavily on the value of $\gamma$. (B) Cross-validation procedure splits library into 75\% training to build the model, 25\% test to evaluate. Errors $E_i$ from each iteration are median-summarized. (C) On a $\log-\log$ scale we plot an example of cross-validated error against $\gamma$. The red dot indicates the minimum.}
    \label{fig:xv}
\end{figure*}

An indispensable part of our surrogate model is the adaptivity of its fit. This is largely mediated through the parameter $\gamma$ which controls smoothness of the model. As an illustration, in Figure~\ref{fig:xv}~(A) we display example scatter plots of estimates against truth for recovering $H_2O$ mole fraction using several different values of $\gamma$. We defer full discussion of the simulation setup and results until Section~\ref{sec:eval}, but note for now that the accuracy of the method strongly depends on $\gamma$. While $\gamma \approx 1\times10^{-3}$ appears most accurate in this particular instance, the optimal value will change from one problem to another. The best value of the smoothness parameter $\gamma$ will also depend significantly on the character of the training library including the total number of spectra and the range over which the parameters are varied. Thus to get a model that works well we need to carefully adapt $\gamma$ to the particular problem and library. We will do this using cross-validation, a technique for tuning parameters that is widely used in the statistics and machine learning literature \parencite{Hastie2009}. The remainder of this section will describe this approach. 

Given a library of spectra, our goal is to choose a best value $\gamma^*$ for the smoothing parameter. The most straight-forward way to do this is to try out several values and pick the one that yields the lowest error. However, given $\gamma$, a significant difficulty is determining how to estimate the error. Ideally, one could look at plot like Figure~\ref{fig:xv}~(A) and choose the value with lowest error. However, in a real application, we will not know the true values for the parameters and thus cannot directly compute accuracy. To get around this, since we do know the true parameters for the library spectra used to train the method, one might instead calculate accuracy of the model in approximating these library spectra. However, choosing $\gamma$ based on this error is not workable as our interpolation condition means that the model will perfectly predict spectra in the library and have an error of exactly zero for the set of library spectra regardless of the choice of $\gamma$. Consequently, this will not give us a sense of how well the given value of $\gamma$ will produce a method that generalizes to new spectra not in the library. Nonetheless, as seen in Figure~\ref{fig:xv}~(A), if we are not careful about our choice of $\gamma$ we are likely to over-fit the model and choose a value for $\gamma$ does not generalize well to new and unseen flow parameters. 

Cross-validation solves these issues by holding back some data from the fitting to use in evaluation. This allows us to estimate how well a method will generalize to new data that the fitted model has not yet seen. The approach works by randomly partitioning the library into two sets: (1) a training set, and (2) a test set. The training set is used to build the model and the test set is used to evaluate this fit. Since this test data was not used to build the model, the calculated accuracy will not be precisely zero and will be a better reflection of how well the model generalizes to new and independent data. Typically, we repeat this procedure several times whereby we randomly partition the data, build the model on the training data and then evaluate on the test data. This repetition helps minimize the effects of the randomization. Subsequently, we summarize the overall error as the mean error across iterations. 

Figure~\ref{fig:xv}~(B) illustrates this approach for our problem. Given a library of spectra and a fixed value of $\gamma$ in each iteration we randomly split the library putting 75\% into the training set (white) and keeping 25\% for the test set (blue). The kernel-based model is then trained on the former and evaluated on the latter. For the $i^{th}$ iteration we calculate the error $E_i$ as the average difference between the Raman intensities across the entire spectral range of the approximated and true spectra in the test set. In our application we do five cross-validation iterations and summarize the overall error for $\gamma$ as the mean across iterations. 

This procedure is repeated over a sequence of potential candidates for $\gamma$ whereby we calculate a cross-validated estimate of our method's performance. We then choose $\gamma^*$ as the value of $\gamma$ with the lowest cross-validated error. This step is illustrated in Figure~\ref{fig:xv}~(C) where we plot an example of the cross-validated error over values of $\gamma$. We further summarize this entire approach to finding $\gamma^*$ in Procedure~\ref{proc:xv}. 

Once we have chosen $\gamma^*$ for our problem, we build the model we will actually use by fitting the model using all 100\% of the library and a smoothing parameter set to $\gamma^*$. While the nested loops of this cross-validated fit seems intricate, the entire fitting procedure typically only takes several seconds to several minutes depending on library size. 

\begin{algorithm}[!ht]
  \caption{Optimal smoothing via cross-validation.}
  \begin{algorithmic}
  \Require{\\
  \begin{itemize}
      \item[$\cdot$] $\Gamma$: a sequence of candidates values for $\gamma$.
      \item[$\cdot$] Library data. 
  \end{itemize}}
    \Ensure{$\gamma^*$: optimal value for smoothing.}
  \For{$\gamma$ in $\Gamma$} 
    \For{$i = 1,\ldots,5$}
    \State 1. Randomly partition library into 
    \State \quad\quad 75\% training and 25\% test.
    \State 2. Build model using training set. 
    \State 3. $E_i =$ error when evaluated on test set.  
    \EndFor 
    \State $E_\gamma = \text{mean}(E_1,\ldots,E_5)$.
  \EndFor 
  \State $\gamma^* = \arg\min_{\gamma\in\Gamma} E_\gamma$. 
  \end{algorithmic}\label{proc:xv}
\end{algorithm}

\section{Application on A Simulated Flow} \label{sec:eval}

We explore the efficacy of this kernel-based model by evaluating its ability to aid in quickly and accurately recovering flow parameters from noisy CARS spectra. In particular, we consider a five-parameter problem representing a theoretical dual-pump CARS experiment probing a $H_2/$air combustion environment, where the unknown the parameters of interest are the gas temperature and mole fractions of the four major species: $N_2$, $H_2$, $O_2$ an $H_2O$. The details of this theoretical dual-pump CARS experiment (i.e., the pump and probe laser frequencies and linewidths) were taken from Ref. \parencite{obyrne2007}. As elaborated upon in Ref. \parencite{obyrne2007}, this particular dual-pump CARS instrument is unique in that it allows for the determination of the absolute mole fractions of the four major species resulting from the combustion of hydrogen and air by exciting both the vibrational Raman transitions of the $N_2$ and $O_2$ Q branch in addition to several pure rotational Raman transitions of $H_2$. Therefore, by neglecting minor species and assuming that water vapor is the only non-resonant specie in the post-combustion environment, the mole fraction of $H_2O$ can also be determined as one minus the sum of the other species' mole fractions. In our application the gas properties input to CARSFT to generate the theoretical spectra are inspired by the equilibrium adiabatic flame temperature calculation for equivalence ratios ranging from $\phi = 0.1$ to $\phi = 4$. Consequently, we let the temperature range between 500 and 2500$K$, and the mole fraction range between 0.25 and 0.85 for $N_2$, 0 and 0.60 for $H_2$, 0 and 0.30 for $O_2$ and 0 to 0.40 for $H_2O$. 

To evaluate performance we generate a ``validation'' set consisting of 250 spectra associated with known parameters. The parameters underlying this set are uniformly sampled over the parameter space with the constraint that the mole fractions are positive and sum to one. 
The validation spectra are created by using CARSFT. 
We add noise to each of these validation spectra and then use numerical optimization along with our model to try and recover the true underlying flow parameters. We then compare these estimates to the known truth to get a measure of accuracy. 
This parameter estimation process is summarized in Procedure~\ref{proc:fit}. 

\begin{algorithm}[!ht]
  \caption{Validation data fitting.}
  \begin{algorithmic}
    \State {\bf Step 1:} Generate a library of $N$ theoretical spectra using CARSFT.
    \State {\bf Step 2:} Fit our model to this library.
    \State {\bf Step 3:} Add random noise to the validation spectra.
    \State {\bf Step 4:} Estimate the validation parameters using numerical optimization.
  \end{algorithmic}\label{proc:fit}
\end{algorithm}

The first step of this procedure uses CARSFT to generate a library to train the model. In section~\ref{sec:acc} we will investigate how the structure of this library (including the number of spectra) affects performance. To ensure fair evaluation, this library used to train the model is independent of the validation spectra used to evaluate the fitting process. 

In the second and third steps of Procedure~\ref{proc:fit} we additionally apply two pre-processing transformations to the spectra and parameters: (1) we $\log$-transform the spectral intensities and (2) we $z$-score the parameters. The $\log$-transformation keeps larger spectral peaks from dominating analysis and helps focus fitting on more subtle lineshape features. Logarithmic and square-root transformation have been used previously in the literature for similar reasons \parencite{greif2021,cutler2011,cutler2015}. The $z$-score transformation subtracts off parameter means and divides by the standard deviation. This is done separately for each of the five parameters and helps standardize the potentially disparate scales of different parameters. For example, un-standardized the temperature ranges between 500 and 2,500$K$ while the $N_2$ mole fraction is physically constrained between zero and one. This standardization keeps parameters with larger scales from dominating the model fitting and parameter estimation routines.

Step three also adds Gaussian noise. This helps our validation spectra better simulate measurements encountered in a real experiment. The noise is added after the $\log$-transformation. Following Cutler~et.~al.~(2011) we model this noise as having zero mean but a standard deviation that scales linearly with spectral intensity \parencite{cutler2011}. Later analysis will explore the effect of the signal-to-noise ratio (SNR) on accuracy. We define the SNR as the ratio of the true spectral intensity to the standard deviation of the noise.

In the final step we use numerical optimization to estimate the parameters for each validation spectrum. These estimates are obtained by finding the parameter values that minimize the sum-of-squares difference between the noisy validation spectrum, denoted $s$, and our model $\hat{r}$ so that our parameter estimate $x^*$ is 
\begin{equation}\label{eqn:min}
    x^* = \arg\min_{x\in \mathcal{X}} ||s - \hat{r}(x)||^2
\end{equation}
where $\mathcal{X}$ is our space of permissible parameters. 
In this paper we do this optimization using the gradient-based sequential least squares programming (SLSP) with the constraint on $\mathcal{X}$ that the mole fractions of the gas mixture sum to one \parencite{dieter94}. Nonetheless, our modeling is flexible and generally optimizer agnostic. Thus other numerical optimization approaches may also be accurate. 

\subsection{Evaluation of Accuracy and Speed}\label{sec:acc}

\begin{figure*}[!ht]
    \centering
    \includegraphics[width=\textwidth]{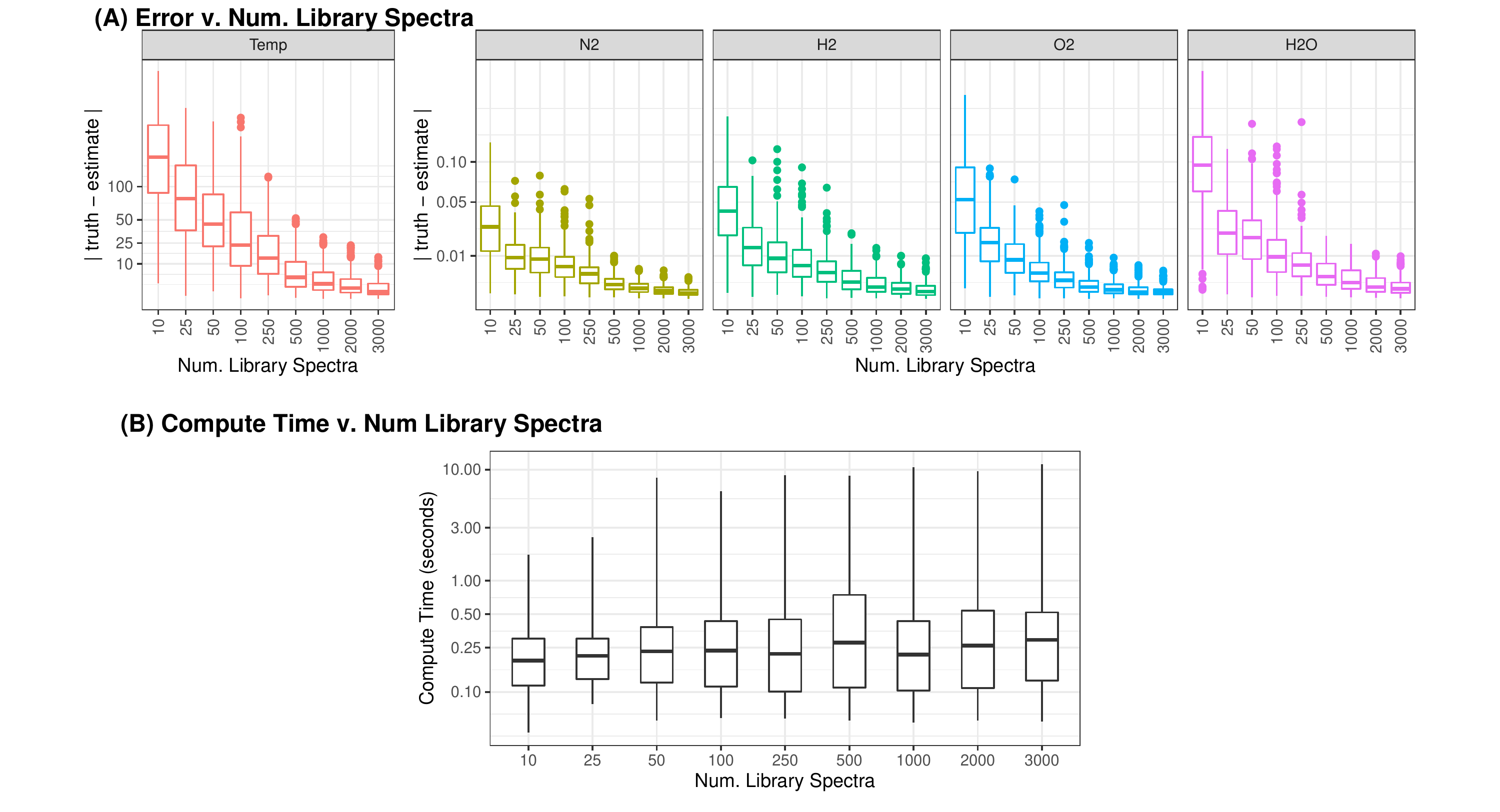}
    \caption{(A) Box-plots of error against number of training spectra in the library for the five parameters. The middle line represents the median and the outer reaches of the box indicate the 25$^{th}$ and 75$^{th}$ percentiles. The length of the box is called the interquartile range (IQR). The whiskers of the box plot indicate the extent of the data beyond the box so long as it is within 1.5 IQRs of the edge of the box. Outlier points beyond this are indicated by individual data points. Here the SNR is fixed at 50. (B) Boxplots of computational time for recovering the validation parameters.}
    \label{fig:ntrain}
\end{figure*}

\begin{figure*}[!ht]
    \centering
    \includegraphics[width=\textwidth]{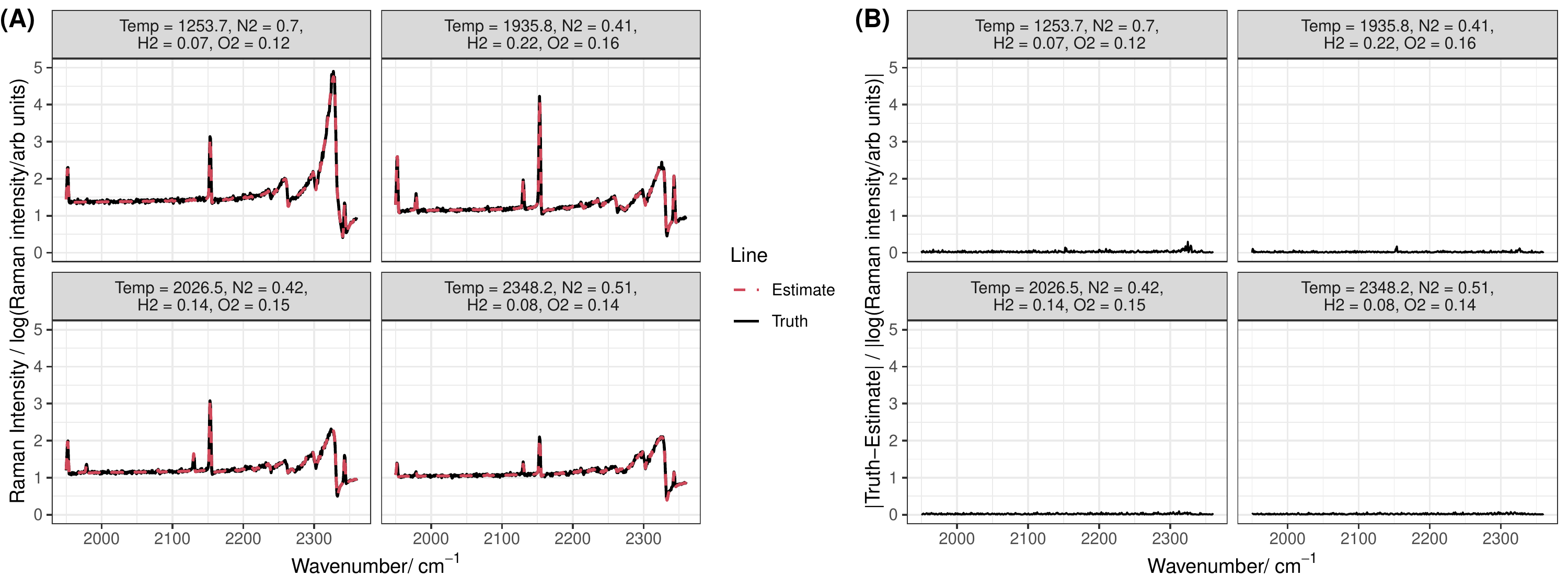}
    \caption{Example best fit approximation spectra and true spectra for four randomly chosen validation spectra using $N=1000$ library spectra. (A) We plot the true spectra (solid black) against the estimate (dashed red). (B) We plot the absolute difference between the true and estimated spectra. }
    \label{fig:ex_fit}
\end{figure*}

\begin{figure*}
    \centering
    \includegraphics[width=\textwidth]{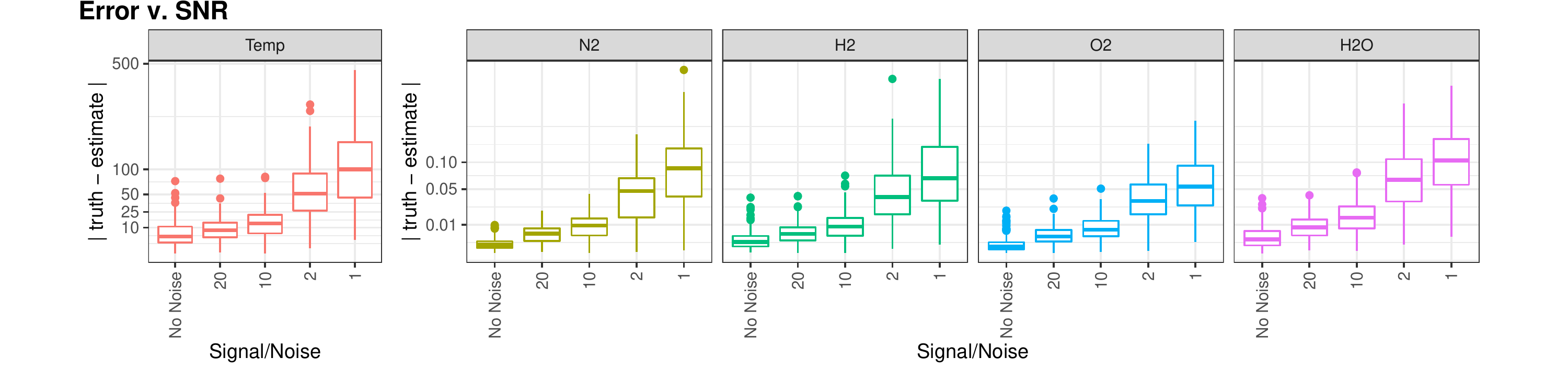}
    \caption{Boxplots of the error for estimated parameters against the truth for temperature and the mole fractions of $N_2$, $H_2$, $O_2$, and $H_2O$. These plots show error for five different signal-to-noise ratios: No noise, 20, 10, 2, and 1. }
    \label{fig:noise}
\end{figure*}

To comprehensively evaluate the accuracy of our adaptively-tuned model we fit the validation data following Procedures~\ref{proc:xv} and \ref{proc:fit} for a range of library sizes and a low amount of noise (SNR of $50$). In Figure~\ref{fig:ntrain}~(A) we display box-plots of error over a sequence of several library sizes ranging over three orders of magnitude. From this figure we can see that given a moderately-sized library of about $1000$ or more spectra our method is able to accurately recover the validation flow parameters. The median absolute error is below $2K$ for temperature, below $0.0008$ for the mole fractions of the resonant species and below $0.0015$ for the non-resonant $H_2O$. Nonetheless, one can increase accuracy by increasing the number of library spectra. In a real problem, the accuracy will depend on a host of considersations like the signal-to-noise ratio of the measured spectra and the structure of the error. We will explore the relationship between SNR and accuracy at the end of this section in Figure~\ref{fig:noise}. However, more broadly, one can tune our approach to achieve the level of error desired. For example, if the estimated error is too high one could increase the number of spectra in the library, or more finely tune $\gamma$ over a wider range of values or number of cross-validation iterations. While Figure~\ref{fig:ntrain} presents an illustrative example of how error behaves as we change the number of library spectra, in practice one will typically want to conduct their own study to determine how many spectra they need to achieve the accuracy they desire based upon the problem's specific experimental setup, expected SNR, number of unknown parameters to be recovered, and speed desired. The adaptable nature of our approach makes such a study easy to do. 

In Figure~\ref{fig:ntrain}~(B) we display box-plots for the time (in seconds) it takes to estimate each of the validation spectra on a 4GHz desktop computer. From this we can see that our approach takes typically about a quarter of a second to fit a spectrum. Note that this figure tallies the compute time (per spectrum) for the final numerical optimization step in Procedure~\ref{proc:fit} not the compute time for $\hat{r}$. Indeed, the numerical optimization typically evaluates this surrogate model $\hat{r}$ hundreds or thousands of times during each fit. Correspondingly, our approach only typically takes around $0.25$ milliseconds for each computation of $\hat{r}$. As a point of comparison, the CARSFT spectrum calculator takes between one and six seconds for a single calculation of $r$ which makes our approach about four orders of magnitude faster. Thus, while it will take our method about a minute to estimate parameters for all 250 validation spectra, if one were to directly replace our $\hat{r}$ with the CARSFT spectrum calculator in Procedure~\ref{proc:fit} it could take a week to do the same calculations.

In Figure~\ref{fig:ex_fit} we plot four randomly selected validation spectra and their best-fit estimated spectrum following Procedure~\ref{proc:fit} using $N=1000$ library spectra. In subplot (A) we overlay the true and estimated spectra and in subplot (B) we plot the absolute difference between the two. We can see from this figure that the best estimated spectra closely follows the true spectra. Note here that our estimates contain error from both (1) the approximation of the underlying truth using the library and (2) noise from the fitting process. Nonetheless, the median difference between truth and estimate is less than 0.02 (on the log-transformed scale).

\begin{figure*}
    \centering
    \includegraphics[width=\textwidth]{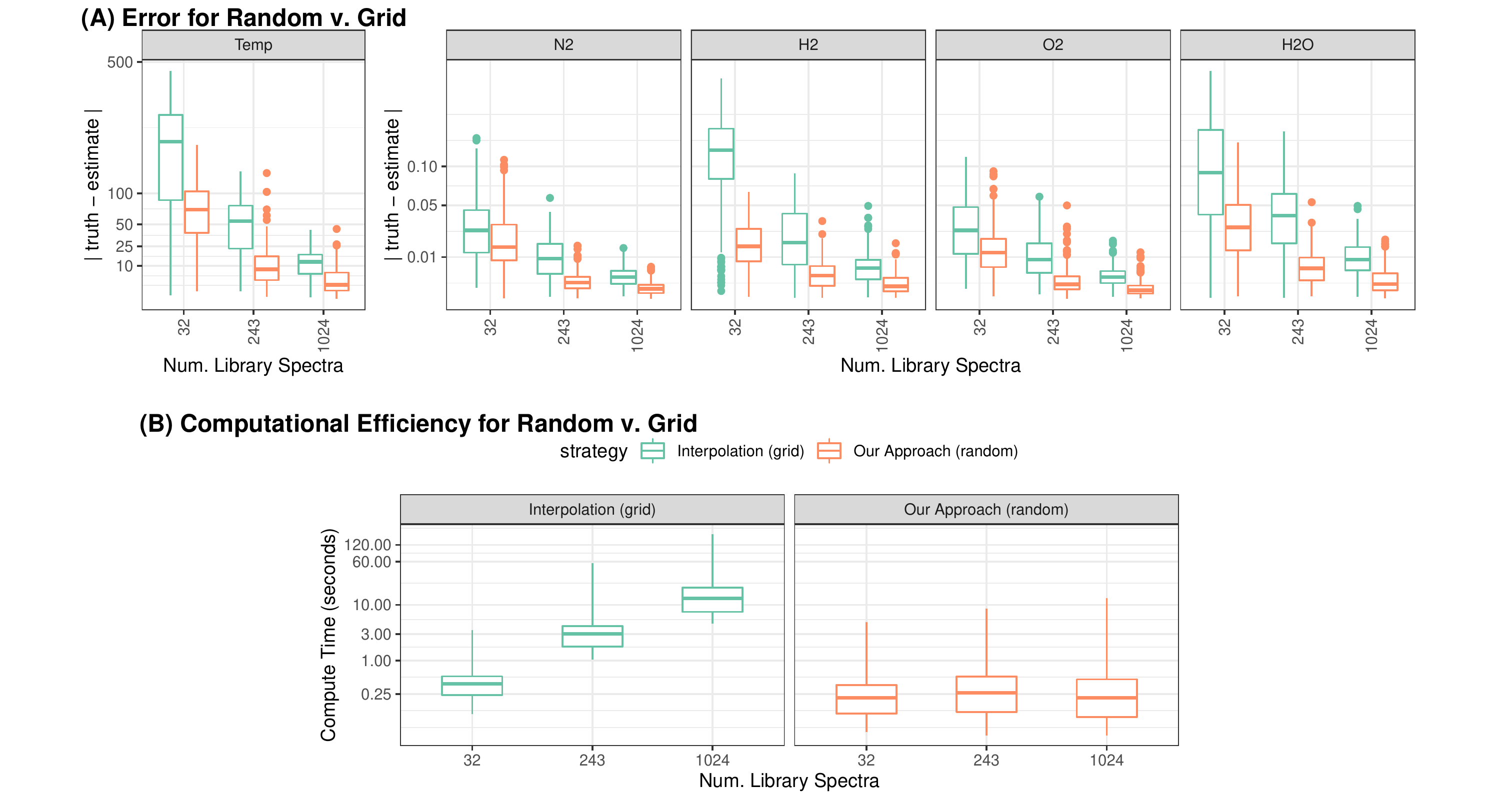}
    \caption{(A) Box-plots of error against number of training spectra for two different strategies: (1) our approach with a randomized library (orange) and (2) a Lagrange interpolation with a regular grid library (green). (C) Computational time of estimating parameters for two strategies. }
    \label{fig:grid}
\end{figure*}

The flexibility of our approach is in stark contrast to popular methods in the literature like Lagrange interpolation. As discussed, such interpolation requires that the library be generated over a regular grid of parameters. This extremely restrictive constraint makes it difficult to navigate the accuracy-speed trade-off. To illustrate this, in Figure~\ref{fig:grid}~(A) we compare accuracy between our approach using a random library of parameters and Lagrange interpolation using a gridded approach. Here, we fit the Lagrange interpolator on three different size grids: (1) using two values for each of the five parameters for a total of $N=2^5 = 32$ library spectra, (2) using three values per parameter so that $N=3^5 = 243$ and (3) using four values per parameter for $N=4^5 =1024$ spectra. The values for temperature in this grid are evenly spaced between $500$ and $2500K$ while the values for each of the species mole fractions are evenly spaced over their ranges. Additionally, we fit our cross-validated kernel-based approach to a sequence similarly-sized libraries. However, the libraries for our approach are randomly sampled uniformly over the same space.

We can see from Figure~\ref{fig:grid}~(A) that our approach typically can produce more accurate estimates using fewer library spectra. The reason for this is that our adaptive approach allows for a library that is more carefully tailored to the specific problem. Our randomized library does not form a regular grid but restricts itself to the physically meaningful parameter space. Conversely, the gridded library cannot enforce this constraint as it does not form a regular grid. Consequently, the gridded Lagrange approach wastes a significant portion of finite library resources interpolating over a nonphysical parameter space that will never be evaluated in the numerical optimization routine. Conversely, our flexible approach is able to pack the same number of library spectra into the important portion of the space and thus learn a more accurate approximation using the same number of spectra.

Our approach can also recover the underlying parameters significantly quicker than Lagrange interpolation as illustrated in Figure~\ref{fig:grid}~(B). While for both methods compute time increases with library size, even using $N=1024$ spectra our approach typically only takes a quarter-second to estimate each parameter while the Lagrange interpolation correspondingly takes around 10 seconds. Thus while it takes our approach about a minute to estimate the 250 validation spectra it takes the Lagrange approach well over an hour.

Finally, in Figure~\ref{fig:noise} we display boxplots plots of the error for the estimated flow parameters for several different signal-to-noise ratios. Here we use our method with a library of $N=500$ spectra varying the SNR from no noise down to a SNR of one. This figure shows that the approach we take is quite accurate until the SNR drops down to a ratio of two or one. Unfortunately, in such high-noise scenarios obtaining accurate estimates becomes significantly difficult with any method and tweaking the library size or numerical optimization parameter is unlikely to yield too much improvement. Nonetheless, in situations where noise is well-controlled recovering multiple parameters from CARS spectra using our kernel-based approach can be done both quickly and accurately. 

\section{Conclusion}\label{sec:conclusion}

In this work we looked at simultaneously recovering several parameters from CARS spectra in the context of a combustion problem. We propose an adaptively tuned kernel-based approach to help efficiently solve this problem. While our examples are from the combustion literature, the method we propose is generally applicable to similar CARS problems in other scientific contexts and with other collections of parameters. For example, the approach could be used to help recover temperature in a classic vibrational $N_2$ CARS experiment. Alternatively, this approach could be used to recover temperature and relative mole fractions of the resonant species if there are multiple non-resonant species whose contributions are unknown. It can also be used to recover the temperature and absolute mole fractions of the resonant species if the susceptibility of the non-resonant species is known. A power of the method we propose is that it is flexible in several ways. For example, it can be extended to an arbitrary number of parameters and does not make any physical assumptions about the spectra or flow regime in its approximation. Consequently it does not have physical restrictions on its applicability beyond the limitations of the experiment itself. Although not directly validated in this work, the generality of our approach means it can likely be used to approximate other classes of spectra like Rayleigh-Brillouin. Indeed, the method we propose is broadly inspired by approximation ideas from that literature \parencite{Hunt2020}. Furthermore, our approach can use simple out-of-the-box optimizers and does not require any specially written optimization code. We fully expect that the method will work well with other optimization approaches like gradient-free genetic algorithms. 

Finally, this work helps demonstrate the power of grid-free approaches. In the context of our problem the freedom of such a grid-free method allows us to build up an approximation only using the library spectra that correspond to the physically meaningful part of the parameter space. In comparison to the gridded methods, this means that we can create a faster method that uses fewer spectra to build an approximator that is just as accurate.  More broadly, such approaches allow us to add spectra to our library in an arbitrary manner. This suggest potential avenues for future work to incorporate such grid-free methods into active learning schemes to optimize library size. 




\section*{Author Contributions}

C. R. G. and G. J. H. conceived the idea; A. D. C. provided the CARSFT code and guidance on generating spectra; G. J. H. wrote the analysis code and simulations and made the plots and figures; C. R. G. guided choice of flow regimes for the simulations; G. J. H. and C. R. G. wrote the manuscript; A. D. C. provided insight and feedback on the manuscript.

\printbibliography 

\end{document}